\documentclass[11pt]{article}

\usepackage[preprint]{acl}

\usepackage{times}
\usepackage{latexsym}
\usepackage[T1]{fontenc}
\usepackage[utf8]{inputenc}
\usepackage{microtype}
\usepackage{inconsolata}
\usepackage{graphicx}

\usepackage{amsmath}
\usepackage{amssymb}
\usepackage{amsthm}
\usepackage{mathtools}

\usepackage{algorithm}
\usepackage{algpseudocode}

\usepackage{booktabs}
\usepackage{multirow}

\usepackage{enumitem}

\usepackage{dblfloatfix}

\newtheorem{theorem}{Theorem}
\newtheorem{proposition}[theorem]{Proposition}
\newtheorem{lemma}[theorem]{Lemma}
\newtheorem{corollary}[theorem]{Corollary}
\newtheorem{assumption}{Assumption}

\theoremstyle{definition}

\newcommand{\caDDTree}{\textsc{CaDDTree}}

\title{Cost-Aware Diffusion Draft Trees for Speculative Decoding}

\author{%
  Shuai Zhang$^{1,2}$ \quad Huachuan Qiu$^{2}$ \quad Hongliang He$^{2}$ \quad Yong Dai$^{2}$ \\
  $^1$Zhejiang University \quad $^2$Westlake University \\
  \texttt{zhangshuai@westlake.edu.cn}
}

\begin{document}
\maketitle

\begin{abstract}
Speculative decoding accelerates inference by having a lightweight drafter propose tokens verified in parallel by the target language model.
Block diffusion drafters such as DFlash generate an entire draft block in one pass, yielding per-position marginals; DDTree uses these to build a candidate tree that maximizes expected acceptance length under a fixed node budget.
We observe, however, that acceptance length is non-decreasing in budget: it always favors larger trees regardless of verification cost, offering no principled basis for budget selection.
We introduce \textbf{CaDDTree} (Cost-aware Diffusion Draft Tree), a method that directly optimizes token throughput (expected tokens generated per unit time) by jointly selecting the tree structure and node budget.
We model draft and verification latencies explicitly, show that the throughput objective decomposes into a per-round one-dimensional search over the budget, and prove that under a convex verification cost the throughput function is \emph{unimodal}, enabling an efficient greedy stopping rule.
CaDDTree requires no offline budget search, adapting the budget each round from the current per-position distributions and verification cost.
Experiments on Qwen3-4B and Qwen3-8B across eight benchmarks spanning reasoning, coding, and instruction-following tasks show that \caDDTree{} matches or surpasses DDTree with oracle budget selection on nearly all tasks.

\end{abstract}

\section{Introduction}
\label{sec:intro}

Autoregressive language models generate text token by token, requiring a full forward pass at every step.
Speculative decoding \citep{leviathan2023fast, chen2023accelerating} addresses this bottleneck by pairing a lightweight drafter with the target language model: the drafter proposes candidate tokens, and the target model verifies them in a single parallel forward pass, preserving the output distribution.

\emph{Tree-based verification} \citep{miao2024specinfer} extends this by organizing multiple candidate continuations as a tree: common prefixes are shared and all branches are verified in a single parallel forward pass.
Each additional node raises coverage but also increases verification cost.

This tension between coverage and cost is the central optimization problem in draft tree construction: \emph{how many nodes should the tree have, and where should they be placed?}
Existing methods \citep{wang2025opt, ringel2026accelerating, li2024eagle2, liu2026dart} optimize the draft tree to maximize expected token acceptance.
While intuitive, this objective is misaligned with throughput: acceptance length is non-decreasing in budget and provides no principled basis for budget selection. Methods that adapt tree shape per step \citep{li2024eagle2, liu2026talon} still fix the total node budget offline, leaving no mechanism to respond to per-round variability in the optimal budget.

We propose throughput as the primary design criterion for draft tree construction.
By modeling drafting and verification costs as functions of tree size, we derive a per-round objective that decomposes into a one-dimensional budget search.
The central finding is that under a mild \emph{convexity condition} on the verification cost, the throughput objective is unimodal in tree size.
This unimodality yields an efficient greedy algorithm that finds the throughput-optimal tree and adapts the budget per round, without any offline budget search or additional training.

We instantiate the framework on block diffusion drafting \citep{arriola2025block, chen2026dflash}, which produces per-position marginal distributions in a single forward pass.
Experiments on Qwen3-4B and Qwen3-8B across eight benchmarks spanning reasoning, coding, and instruction-following tasks show that \caDDTree{} matches or surpasses DDTree with oracle budget selection on nearly all tasks.
Code is available at \url{https://github.com/ZhangShuai1230/CaDDTree}.

\section{Related Work}
\label{sec:related}

\paragraph{Speculative decoding and drafter design.}
Speculative decoding \citep{leviathan2023fast, chen2023accelerating} uses a cheap drafter to propose tokens verified in one parallel forward pass by the target model, preserving its distribution.
Autoregressive drafters conditioned on target hidden states, most notably the EAGLE family \citep{li2024eagle, li2024eagle2, li2026eagle}, achieve high acceptance rates.
Block diffusion models \citep{arriola2025block} offer a complementary paradigm: a single forward pass predicts an entire token block, enabling longer drafts at near-constant latency.
DFlash \citep{chen2026dflash} instantiates this by conditioning the drafter on target-model features, achieving state-of-the-art performance.
PARD \citep{an2025pard} adapts autoregressive models to mimic block-parallel prediction.

\paragraph{Tree-based parallel verification.}
SpecInfer \citep{miao2024specinfer} introduced tree attention, enabling verification of an entire candidate tree in one forward pass.
Medusa \citep{cai2024medusa} exploits this with lightweight prediction heads covering successive future positions.
EAGLE-2 \citep{li2024eagle2} constructs draft trees dynamically guided by per-step acceptance-rate estimates; EAGLE-3 \citep{li2026eagle} further improves draft quality by fusing multi-layer features.
DySpec \citep{xiong2025dyspec}, Staged Speculative Decoding \citep{spector2023accelerating}, and Recursive Speculative Decoding \citep{jeon2024recursive} explore further structural extensions.
DART \citep{liu2026dart} builds draft trees from one-pass parallel logits with $n$-gram continuity constraints.
TALON \citep{liu2026talon} adapts tree shape per step using token confidence; GOOSE \citep{jin2026goose} builds anisotropic trees routing reliable tokens into chains and uncertain ones into wide branches. Both fix the total node budget offline.

\paragraph{Optimal tree construction.}
OPT-Tree \citep{wang2025opt} first formulated adaptive tree construction for autoregressive drafters, maximizing expected acceptance length under a fixed node budget.
DDTree \citep{ringel2026accelerating} extends this to block diffusion drafters: one pass yields per-position marginals for greedy optimal tree assembly.
Zhang et al. \citep{zhang2026learning} optimize throughput directly via reinforcement learning but require training additional draft policies.
CaDDTree departs from all of the above: it replaces acceptance-length maximization with direct throughput optimization, adapts the total node budget per round, and provably identifies the optimum via a greedy algorithm with unimodality guarantees, requiring no additional training beyond a one-time cost profiling step.

\section{Background}
\label{sec:background}

\begin{figure*}[tp]
\centering
\begin{minipage}{0.48\textwidth}
  \centering
  \includegraphics[width=\textwidth]{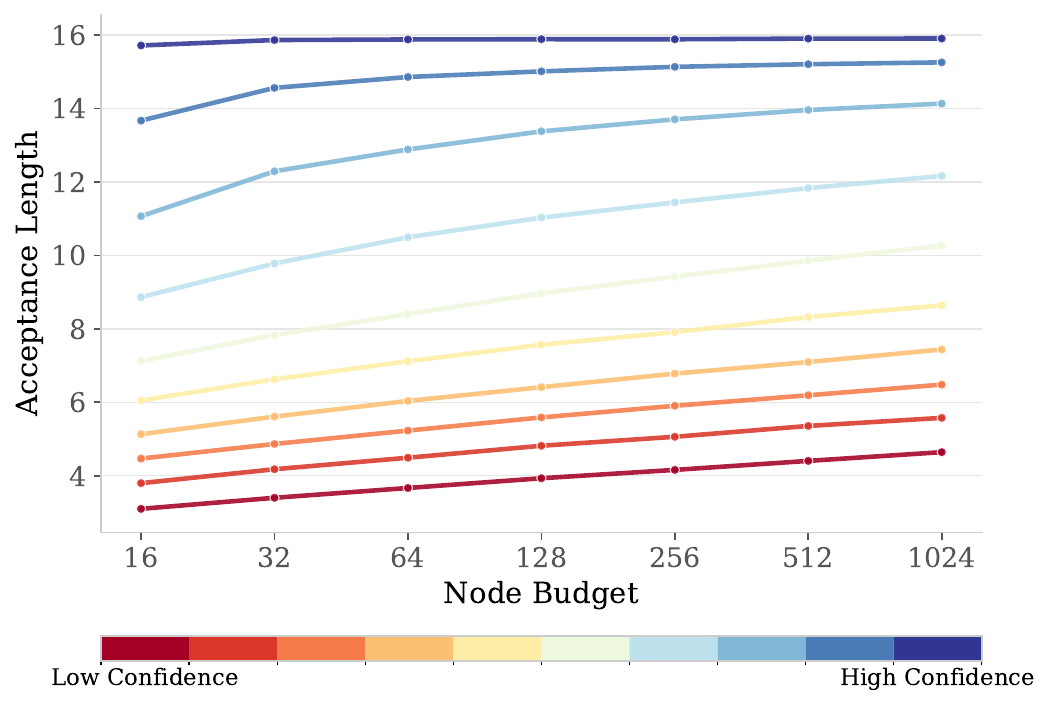}
\end{minipage}
\hfill
\begin{minipage}{0.48\textwidth}
  \centering
  \includegraphics[width=\textwidth]{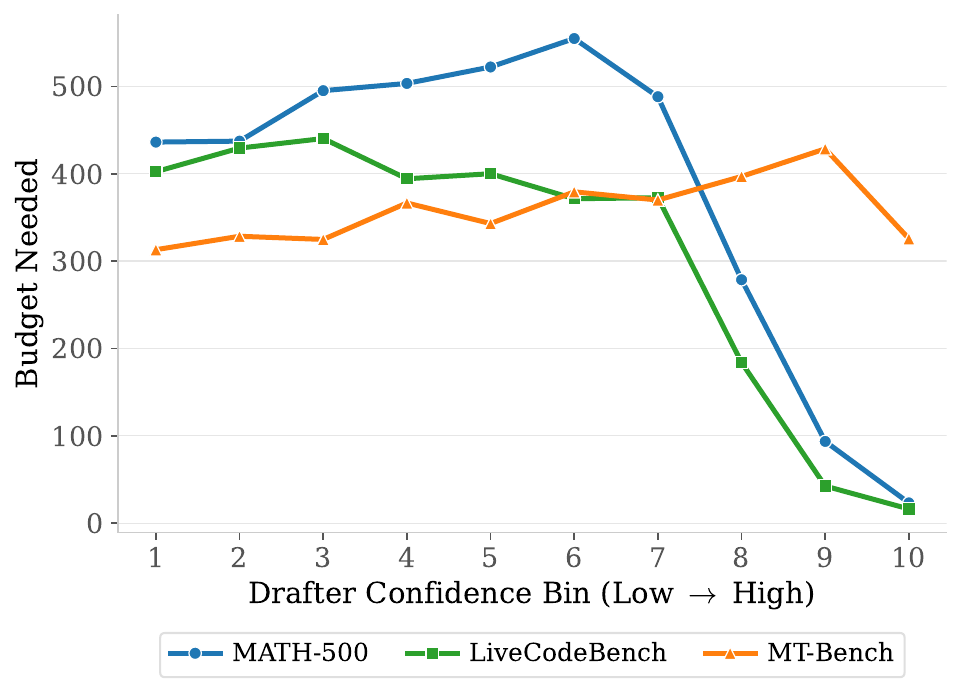}
\end{minipage}
\caption{\emph{Left}: Acceptance length vs.\ node budget for ten drafter-confidence deciles (Bin~1: low, Bin~10: high), Qwen3-4B on MATH-500 at temperature~0.0. High-confidence rounds saturate at small budgets; low-confidence rounds continue to benefit from larger trees throughout the range.
\emph{Right}: Budget required to recover 100\% of the 1024-node acceptance length across confidence deciles for three datasets (Qwen3-4B, temperature~0.0). All decrease with confidence; reasoning tasks drop sharply while MT-Bench decreases later and more gradually.}
\label{fig:confidence-analysis}
\end{figure*}

\subsection{Speculative Decoding}
\label{sec:bg-sd}

Let $p(\cdot \mid \mathbf{c})$ denote the target model's distribution over the next token given context $\mathbf{c}$, and let $q(\cdot \mid \mathbf{c})$ denote the drafter's distribution.
Speculative decoding \citep{leviathan2023fast, chen2023accelerating} accelerates generation by having the drafter propose a sequence of $n$ candidate tokens $(x_1, \ldots, x_n)$, then letting the target model verify all of them in a single parallel forward pass.
For each position $i$, the target evaluates $p(x_i \mid \mathbf{c}, x_{<i})$ and accepts or rejects via rejection sampling \citep{leviathan2023fast}: accepted tokens extend the output, while the first rejection triggers a resample from a corrected distribution, preserving the target's distribution exactly.
Under greedy decoding, acceptance is equivalent to requiring $x_i = \arg\max p(\cdot \mid \mathbf{c}, x_{<i})$.

Each round begins with one committed token $b$, the \emph{bonus token}: it comes from the prefill pass in the first round and from the last accepted token in subsequent rounds.
Let $\alpha \ge 0$ denote the number of tokens accepted in a round; the round commits $1 + \alpha$ tokens in total.
Throughput scales with $\mathbb{E}[\alpha]$, making draft quality the central design objective.

\subsection{Block Diffusion Drafting}
\label{sec:bg-dllm}

Autoregressive drafters require one forward pass per drafted token.
Block diffusion models \citep{arriola2025block} eliminate this cost: a single forward pass given context $\mathbf{c}$ and bonus token $b$ produces logits for $L$ positions simultaneously, yielding a per-position \emph{marginal distribution}
\begin{equation}
  q_i(v \mid \mathbf{c}, b) = \operatorname{softmax}(\ell_i)_v,
  \label{eq:marginals}
\end{equation}
for $v \in \mathcal{V}$ and $i = 1, \ldots, L$.
Each $q_i$ is a \emph{marginal}: it conditions only on $(\mathbf{c}, b)$, not on tokens chosen at earlier positions within the same block, so the drafter's joint distribution factorizes as
\begin{equation}
  Q(y_{1:L} \mid \mathbf{c}, b) := \prod_{i=1}^{L} q_i(y_i \mid \mathbf{c}, b).
  \label{eq:factorized}
\end{equation}
This one-pass, fixed-cost structure makes long draft sequences feasible regardless of depth; the trade-off is that positions are predicted independently, which can accelerate error accumulation.
DFlash \citep{chen2026dflash} instantiates block diffusion conditioned on target-model features, achieving state-of-the-art acceptance rates.

\subsection{Tree Drafting and Tree Attention}
\label{sec:bg-tree}

Rather than proposing a single continuation, the drafter can propose a \emph{tree} of candidates verified simultaneously in one target-model forward pass via \emph{tree attention} \citep{miao2024specinfer}.
Covering multiple branches increases the probability that the target's preferred continuation appears somewhere in the draft.

Formally, a \emph{draft tree} $\mathcal{T}$ is a set of nodes, where each node $\mathbf{s} = (s_1, \ldots, s_d)$ represents the depth-$d$ token sequence $(b, s_1, \ldots, s_d)$; $\mathcal{T}$ is \emph{valid} if it is prefix-closed: whenever $\mathbf{s} \in \mathcal{T}$, every prefix of $\mathbf{s}$ also belongs to $\mathcal{T}$.
In tree attention, each node attends only to the shared context KV cache and to its own ancestors.
The target model selects the path with the longest accepted prefix; that prefix is appended to the output, and the first unaccepted token along the chosen path becomes the next bonus token.

\subsection{Draft Tree Construction}
\label{sec:bg-tree-construction}

Given a draft tree $\mathcal{T}$, the ideal objective is to maximize the expected acceptance length under the target distribution $p$:
\begin{equation}
  \max_{\mathcal{T}} \; \mathbb{E}_p\bigl[\alpha(\mathcal{T}, \mathbf{c})\bigr],
  \label{eq:true-obj}
\end{equation}
where $\alpha(\mathcal{T}, \mathbf{c})$ denotes the number of tokens accepted.
This is intractable: $p$ is the expensive target model itself, and evaluating $\mathbb{E}_p[\alpha]$ requires marginalizing over all paths in $\mathcal{T}$ under $p$.

\citet{ringel2026accelerating} propose to replace $p$ by the drafter's factorized distribution $Q$ from Equation~\eqref{eq:factorized}, yielding a tractable surrogate.
Under $Q$, the expected number of accepted tokens decomposes over nodes as
\begin{equation}
  \Phi(\mathcal{T}) := \sum_{\mathbf{s} \in \mathcal{T}} \pi_{\mathbf{s}},
  \label{eq:surrogate}
\end{equation}
where the \emph{prefix probability}
\begin{equation}
  \pi_{\mathbf{s}} := \prod_{i=1}^{d} q_i(s_i \mid \mathbf{c}, b)
  \label{eq:prefix-prob}
\end{equation}
is the probability that a sample from $Q$ begins with $\mathbf{s}$.
The factorization of $Q$ is what makes $\Phi(\mathcal{T})$ computable from a single drafter forward pass.

Optimizing $\Phi(\mathcal{T})$ under a fixed budget $n = |\mathcal{T}|$ reduces to selecting the $n$ nodes with the largest prefix probabilities.
Labeling all prefix probabilities in non-increasing order as $\delta_1 \ge \delta_2 \ge \cdots$,
\begin{equation}
  \Phi^*(n) := \max_{|\mathcal{T}|=n} \Phi(\mathcal{T}) = \sum_{k=1}^{n} \delta_k,
  \label{eq:phistar}
\end{equation}
with the optimal tree $\mathcal{T}^*_n$ consisting of the top-$n$ nodes by $\pi_\mathbf{s}$.
DDTree constructs $\mathcal{T}^*_n$ greedily: a max-heap over log-prefix-scores pops nodes in order of decreasing $\delta_k$, pushing the first child and next sibling at each step, and halts after $B$ pops.
The tree size $B$ is a hyperparameter with no principled criterion for setting it.


\begin{figure}[tbp]
\centering
\includegraphics[width=\columnwidth]{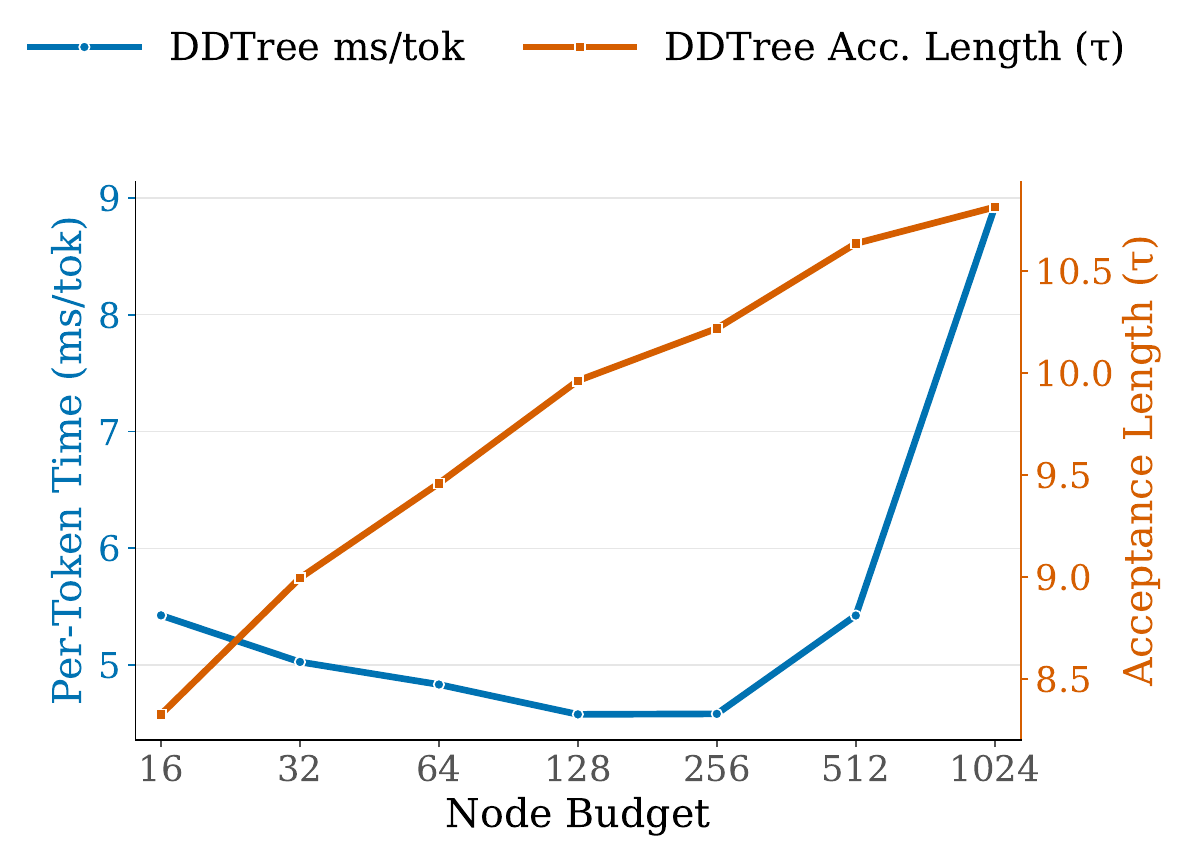}
\caption{Acceptance length $\tau$ (right axis) and per-token generation time (left axis) vs.\ node budget, for Qwen3-4B on MATH-500 at temperature~0.0. Acceptance length is monotone; per-token time is unimodal and minimized at a budget well below the acceptance-length maximum.}
\label{fig:time-vs-budget}
\end{figure}

\begin{table*}[tp]
\centering
\small
\setlength{\tabcolsep}{4pt}
\caption{Per-token generation time (ms/tok, lower is better) and mean acceptance length $\tau$ at temperature~0.0. $n^*$ is the oracle's best grid budget; $\bar{n}^*(\sigma)$ is \caDDTree{}'s per-round mean (std). \caDDTree{} matches or surpasses DDTree-oracle on nearly all tasks while selecting its budget automatically each round.}
\label{tab:main-results}
\begin{tabular*}{\linewidth}{@{\extracolsep{\fill}}l c cc ccc ccc}
\toprule
 & \textbf{AR}
 & \multicolumn{2}{c}{\textbf{DFlash}}
 & \multicolumn{3}{c}{\textbf{DDTree-oracle}}
 & \multicolumn{3}{c}{\textbf{\caDDTree{} (ours)}} \\
\cmidrule(lr){3-4}\cmidrule(lr){5-7}\cmidrule(lr){8-10}
\textbf{Dataset}
  & ms/tok
  & ms/tok & $\tau$
  & ms/tok & $\tau$ & $n^*$
  & ms/tok & $\tau$ & $\bar{n}^*(\sigma)$ \\
\midrule
\multicolumn{10}{c}{\emph{Qwen3-4B}} \\
MATH-500       & 30.65 & 5.93 & 7.71 & \textbf{4.53} & 10.35 & 256 & \textbf{4.53} & 10.37 & $221\,(57)$ \\
GSM8K          & 31.11 & 6.96 & 6.53 & \textbf{5.14} & 9.08  & 256 & 5.18          & 9.13  & $249\,(61)$ \\
AIME 2025      & 31.64 & 6.28 & 7.43 & \textbf{4.95} & 9.40  & 128 & 4.96          & 9.64  & $215\,(37)$ \\
HumanEval      & 30.87 & 6.82 & 6.64 & \textbf{4.97} & 9.32  & 256 & 5.01          & 9.32  & $224\,(59)$ \\
MBPP           & 30.68 & 7.41 & 6.14 & \textbf{5.26} & 8.87  & 256 & 5.31          & 8.91  & $257\,(59)$ \\
LiveCodeBench  & 30.82 & 6.57 & 7.09 & 4.97          & 9.31  & 128 & \textbf{4.93} & 9.64  & $197\,(56)$ \\
MT-Bench       & 31.07 & 12.59 & 4.39 & 8.34         & 6.47  & 256 & \textbf{8.32} & 6.44 & $235\,(37)$ \\
Alpaca         & 31.08 & 16.34 & 3.12 & \textbf{10.42} & 4.89 & 256 & 10.66        & 4.88 & $264\,(41)$ \\
\midrule
\multicolumn{10}{c}{\emph{Qwen3-8B}} \\
MATH-500       & 30.95 & 5.83 & 7.91 & \textbf{4.58} & 10.20 & 128 & 4.60          & 10.32 & $165\,(42)$ \\
GSM8K          & 30.87 & 6.99 & 6.51 & 5.22          & 8.81  & 128 & \textbf{5.19} & 9.04  & $184\,(24)$ \\
AIME 2025      & 31.44 & 6.11 & 7.49 & 5.13          & 9.18  & 128 & \textbf{5.00} & 9.51  & $156\,(40)$ \\
HumanEval      & 31.10 & 7.00 & 6.48 & 5.14          & 8.98  & 128 & \textbf{5.11} & 9.20  & $172\,(38)$ \\
MBPP           & 31.01 & 7.75 & 5.93 & \textbf{5.53} & 8.42  & 128 & \textbf{5.53} & 8.61  & $187\,(20)$ \\
LiveCodeBench  & 30.66 & 6.56 & 7.17 & \textbf{5.03} & 9.37  & 128 & 5.04          & 9.53  & $152\,(49)$ \\
MT-Bench       & 30.38 & 12.71 & 4.28 & 8.59         & 6.10  & 128 & \textbf{8.58} & 6.22  & $172\,(25)$ \\
Alpaca         & 30.14 & 15.78 & 3.08 & 10.43        & 4.68  & 128 & \textbf{10.34} & 4.80 & $186\,(11)$ \\
\bottomrule
\end{tabular*}
\end{table*}

\section{Throughput-Optimal Draft Tree Construction}
\label{sec:method}

\subsection{The Fixed-Budget Trap}
\label{sec:method-problem}

The DDTree framework frames tree construction as: given a budget $B$, build $\mathcal{T}^*_B$.
This leaves the budget as a free hyperparameter, and there is no principled criterion for setting it.
Three empirical observations motivate replacing the acceptance-length objective with direct throughput maximization over an adaptive, context-aware budget.

\paragraph{Observation 1: per-token time is unimodal, not monotone.}
Figure~\ref{fig:time-vs-budget} plots acceptance length and per-token generation time as functions of the node budget for Qwen3-4B on MATH-500.
Acceptance length increases monotonically: more nodes give the target model a larger set of candidates to accept.
Per-token time, however, is unimodal: it decreases initially because larger trees yield more accepted tokens per round, making the fixed drafting cost cheaper per output token, then increases once the marginal verification overhead outweighs the additional acceptance gain.
The throughput-optimal budget is strictly smaller than the acceptance-length-maximizing budget: the latter always selects a larger tree than throughput can justify.

\paragraph{Observation 2: the optimal budget varies from round to round.}
The optimal budget is not a property of the dataset alone; it varies at each decoding round depending on the drafter's confidence.
Figure~\ref{fig:confidence-analysis} quantifies this using Qwen3-4B on MATH-500, grouping rounds into ten confidence deciles by their maximum position-1 probability.

The left panel shows acceptance length as a function of budget per decile.
High-confidence rounds saturate quickly (further nodes add negligible acceptance gain), while low-confidence rounds continue to benefit from larger trees throughout the budget range.
The right panel shows that the budget needed for full acceptance recovery varies substantially with drafter confidence across all three datasets; reasoning tasks saturate earlier and more sharply than MT-Bench.
No single fixed budget is efficient across all confidence levels and tasks.

\begin{figure}[tbp]
\centering
\includegraphics[width=\columnwidth]{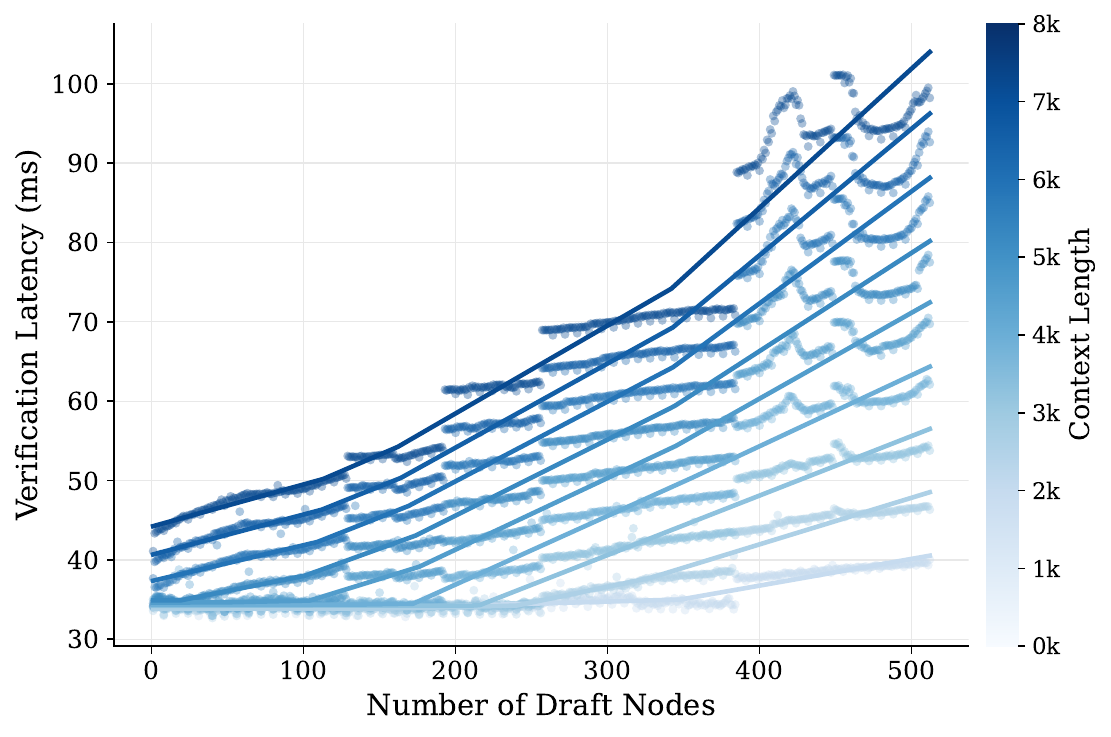}
\caption{Verification latency vs.\ number of draft nodes for Qwen3-4B on 8$\times$A800 across nine context lengths (0--8k tokens, light to dark).
Dots are raw measurements; curves are convex regression fits.
Latency is monotone and convex in node count at every context length; longer contexts shift the curve upward and increase its slope.}
\label{fig:cost-profile}
\end{figure}

\paragraph{Observation 3: verification cost grows with tree size and context.}
Figure~\ref{fig:cost-profile} shows that verification latency increases with node count, and becomes more expensive at longer context lengths.
This matters because context length grows throughout generation, so verification cost keeps changing even within a single response.

\subsection{Throughput as the Optimization Objective}
\label{sec:method-objective}

The observations above suggest that per-token latency, rather than acceptance length, should be the design criterion.

\paragraph{Cost model.}
Let $\ell$ denote the current context length.
Define the round cost $C(\mathcal{T};\,\ell) := C_d + C_v(|\mathcal{T}|;\,\ell)$, where $C_d > 0$ is the drafting cost and $C_v(n;\,\ell)$ is the verification cost for a tree with $n$ nodes at context length $\ell$.
We model $C_v$ as a function of $n$ and $\ell$; Section~\ref{sec:exp-cost-model} validates the resulting cost estimates.

\paragraph{Per-round objective.}
Let $\mathcal{T}_t$ denote the draft tree built at round $t$ and $\alpha_t$ the number of tokens accepted.
The response throughput over $R$ rounds is $\Theta_R = \sum_t (1+\alpha_t) / \sum_t C(\mathcal{T}_t;\,\ell_t)$.
Under the surrogate approximation (derivation in Appendix~\ref{sec:proof-throughput}), maximizing long-run throughput reduces round-by-round to maximizing
\begin{equation}
  \hat{\theta}(\mathcal{T};\,\ell) := \frac{1 + \Phi(\mathcal{T})}{C_d + C_v(|\mathcal{T}|;\,\ell)}.
  \label{eq:per-round}
\end{equation}

\subsection{Decomposition into Budget Selection}
\label{sec:method-decompose}

The objective $\hat{\theta}(\mathcal{T};\,\ell)$ ranges over all valid trees.
We first reduce this combinatorial problem to a search over a single integer.

\begin{proposition}[Structure-Budget Decomposition]
\label{prop:decomposition}
For any $n \ge 0$, the valid tree with exactly $n$ nodes maximizing $\hat{\theta}(\cdot;\,\ell)$ is the greedy best-first tree $\mathcal{T}^*_n$.
Consequently,
\begin{align}
  \max_{\mathcal{T}\;\text{valid}} \hat{\theta}(\mathcal{T};\,\ell)
    &= \max_{n \ge 0} \hat{\theta}^*(n;\,\ell), \label{eq:decomposition} \\
  \hat{\theta}^*(n;\,\ell) &:= \frac{1 + \Phi^*(n)}{C_d + C_v(n;\,\ell)}. \nonumber
\end{align}
\end{proposition}

Proposition~\ref{prop:decomposition} reduces the problem (proof: Appendix~\ref{sec:proof-decomp}) to the one-dimensional search $\max_{n \ge 0} \hat{\theta}^*(n;\,\ell)$, evaluated incrementally via the same best-first heap used to build the DDTree, with $\delta_n = \Phi^*(n) - \Phi^*(n-1)$ being the $n$-th largest prefix probability.

\subsection{Unimodality of Surrogate Throughput}
\label{sec:method-unimodal}

All proofs in this section are in Appendix~\ref{sec:proofs}.
We identify a general condition under which $\max_n \hat{\theta}^*(n;\,\ell)$ is efficiently solvable.
Define the \emph{marginal verification cost} $c_n := C_v(n;\,\ell) - C_v(n-1;\,\ell)$ (for fixed $\ell$) and the
\emph{marginal acceptance-cost ratio}
\begin{equation}
  \rho_n := \frac{\delta_n}{c_n}.
  \label{eq:marginal-ratio}
\end{equation}

\begin{assumption}[Diminishing Benefit-Cost Ratio]
\label{ass:convex}
The marginal benefit-cost ratios are non-increasing: $\rho_1 \ge \rho_2 \ge \cdots$.
\end{assumption}

\begin{lemma}[Weighted Average Representation]
\label{lem:weighted-avg}
For any $n \ge 0$ with $c_{n+1} > 0$,
\begin{equation}
  \hat{\theta}^*(n+1) = w_n \,\hat{\theta}^*(n) + (1-w_n)\,\rho_{n+1},
  \label{eq:weighted-avg}
\end{equation}
where $w_n = \frac{C_d + C_v(n;\,\ell)}{C_d + C_v(n+1;\,\ell)} \in (0,1)$.
Consequently,
\begin{equation}
  \hat{\theta}^*(n+1) > \hat{\theta}^*(n)
  \;\Longleftrightarrow\;
  \rho_{n+1} > \hat{\theta}^*(n).
  \label{eq:stopping}
\end{equation}
\end{lemma}

\begin{theorem}[Unimodality of Surrogate Throughput]
\label{thm:unimodal}
Under Assumption~\ref{ass:convex}, $\{\hat{\theta}^*(n)\}_{n \ge 0}$ is unimodal: there exists
$n^* \ge 0$ such that $\hat{\theta}^*(n)$ is non-decreasing for $n \le n^*$ and non-increasing
for $n \ge n^*$.
\end{theorem}

\begin{corollary}[Greedy Optimality]
\label{cor:greedy}
The first index $n^*$ satisfying $\hat{\theta}^*(n^*+1) \le \hat{\theta}^*(n^*)$ is the global
maximizer of $\hat{\theta}^*$, and $\mathcal{T}^*_{n^*}$ is the throughput-optimal tree.
\end{corollary}

\begin{corollary}[Convex cost]
\label{cor:convex}
If $C_v(\cdot;\,\ell)$ is convex in $n$ for fixed $\ell$ (marginal cost $c_n$ non-decreasing), Assumption~\ref{ass:convex} holds.
\end{corollary}

\subsection{The \caDDTree{} Algorithm}
\label{sec:method-algorithm}

Algorithm~\ref{alg:caddtree} implements Corollary~\ref{cor:greedy} for any verification cost $C_v(\cdot;\,\ell)$ satisfying Assumption~\ref{ass:convex} at the current context length $\ell$.
It runs a max-heap over log-prefix-probabilities and halts at the first node that fails the stopping condition~\eqref{eq:stopping}.
The tree size $n^*$ adapts jointly to the drafter's confidence distribution and the context-dependent cost, without any hyperparameter.

\begin{algorithm}[tbp]
\caption{\caDDTree{}: Cost-Aware Diffusion Draft Tree Construction}
\label{alg:caddtree}
\begin{algorithmic}[1]
\Require Per-position distributions $\{q_i\}_{i=1}^{L}$ with top-$K$ tokens
         $\{v_i^{(k)}\}$ and log-probs $\{\log q_i^{(k)}\}$;
         current context length $\ell$;
         drafting cost $C_d$; verification cost function $C_v(\cdot\,;\,\ell)$
\Ensure Throughput-optimal draft tree $\mathcal{T}^*$
\State $\mathcal{T}^* \leftarrow \emptyset$,\;
       $\Phi \leftarrow 0$,\;
       $n \leftarrow 0$
\State $\hat{\theta}_{\mathrm{prev}} \leftarrow 1/(C_d + C_v(0;\,\ell))$
\State Initialize max-heap $H$ with $\bigl((1),\;\log q_1^{(1)}\bigr)$
\While{$H \ne \emptyset$}
  \State Pop $\mathbf{s} = (s_1, \ldots, s_d)$ with
         $\sigma(\mathbf{s}) = \sum_{i=1}^d \log q_i^{(s_i)}$
  \State $\delta \leftarrow \exp(\sigma(\mathbf{s}))$
  \State $\hat{\theta}_{\mathrm{new}} \leftarrow
         (1 + \Phi + \delta)\,/\,(C_d + C_v(n+1;\,\ell))$
  \If{$\hat{\theta}_{\mathrm{new}} \le \hat{\theta}_{\mathrm{prev}}$}
    \State \textbf{break}
  \EndIf
  \State Add $(v_1^{(s_1)}, \ldots, v_d^{(s_d)})$ to $\mathcal{T}^*$
  \State $\Phi \mathrel{+}= \delta$;\;
         $n \mathrel{+}= 1$;\;
         $\hat{\theta}_{\mathrm{prev}} \leftarrow \hat{\theta}_{\mathrm{new}}$
  \If{$s_d + 1 \le K$}
    \State Push sibling $(s_1,\ldots,s_{d-1},s_d{+}1)$ with key
           $\sigma(\mathbf{s}) - \log q_d^{(s_d)} + \log q_d^{(s_d+1)}$
  \EndIf
  \If{$d < L$}
    \State Push first child $(s_1,\ldots,s_d,1)$ with key
           $\sigma(\mathbf{s}) + \log q_{d+1}^{(1)}$
  \EndIf
\EndWhile
\State \Return $\mathcal{T}^*$
\end{algorithmic}
\end{algorithm}

\subsection{Offline Cost Profiling}
\label{sec:method-cost-estimation}

\paragraph{Profiling procedure.}
$C_v(n;\,\ell)$ is estimated once offline per (model, hardware) pair.
We profile at a grid of context lengths $\mathcal{L} = \{0,\,1\text{k},\,2\text{k},\,\ldots,\,8\text{k}\}$ tokens and node counts $\mathcal{N} = \{1, 2, \ldots, N\}$ with $N = 1024$.
At each $(\ell, n)$, we run the target model with a dummy tree of $n$ nodes and record the median verification latency over 20 trials (after 5 warm-up iterations).
The drafting cost $C_d$ is measured as the mean latency of a single drafter forward pass over 500 rounds after 5 warm-up iterations.

\paragraph{Convex regression.}
For each $L_j \in \mathcal{L}$, we fit $\hat{C}_v(\cdot;\,L_j)$ by solving
\begin{equation}
  \min_{f \;\text{convex, non-decreasing}} \;\sum_{n\in\mathcal{N}} \bigl(y_{j,n} - f(n)\bigr)^2,
  \label{eq:convex-reg}
\end{equation}
a convex quadratic program with $O(N)$ variables and $O(N)$ constraints, solved in under one second per context length via CLARABEL~\citep{goulart2026clarabel}.
The result is a piecewise-linear function that is convex and non-decreasing by construction, with breakpoints automatically determined by the data.

\paragraph{Context interpolation at inference.}
At run time, the current context length $\ell$ is known exactly.
For $L_j \le \ell < L_{j+1}$, let $\alpha = (\ell - L_j)/(L_{j+1} - L_j)$; we set
\begin{equation}
\begin{split}
  \hat{C}_v(n;\,\ell) &= (1-\alpha)\,\hat{C}_v(n;\,L_j) \\
                       &\quad+ \alpha\,\hat{C}_v(n;\,L_{j+1}).
\end{split}
\label{eq:cv-interp}
\end{equation}
Since linear interpolation of convex functions is convex, $\hat{C}_v(\cdot;\,\ell)$ inherits convexity and monotonicity in $n$, so Assumption~\ref{ass:convex} holds for every $\ell$.


\begin{figure*}[tp]
\centering
\includegraphics[width=\textwidth]{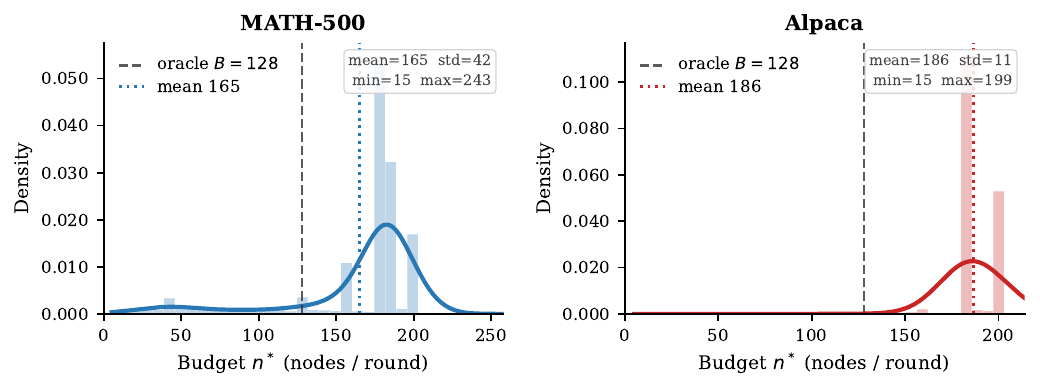}
\caption{Per-round budget $n^*$ distribution for \caDDTree{} on MATH-500 and Alpaca (Qwen3-8B, temp.~0.0). On MATH-500, the budget varies widely (std\,$\approx$\,42), spanning both below and above the grid oracle ($B\!=\!128$). On Alpaca, the distribution is tightly concentrated (std\,$\approx$\,11) near 186, consistently above the oracle. In both cases \caDDTree{} selects $n^*$ automatically each round without any manual tuning.}
\label{fig:adaptive-budget}
\end{figure*}

\section{Experiments}
\label{sec:experiments}

\subsection{Experimental Setup}
\label{sec:exp-setup}

\paragraph{Models.}
We evaluate on two target models: Qwen3-4B and Qwen3-8B \citep{yang2025qwen3}, each paired with its DFlash checkpoint.

\paragraph{Benchmarks.}
Eight benchmarks spanning reasoning, code, and instruction following: MATH-500 \citep{lightman2024let}, GSM8K \citep{cobbe2021training}, AIME 2025, HumanEval \citep{chen2021evaluating}, MBPP \citep{austin2021program}, LiveCodeBench \citep{jain2025livecodebench}, MT-Bench \citep{zheng2023judging}, and Alpaca \citep{taori2023alpaca}.

\paragraph{Baselines.}
\begin{itemize}[leftmargin=*, itemsep=1pt, topsep=2pt]
  \item \textbf{AR}: standard greedy/sampling without speculative decoding.
  \item \textbf{DFlash} \citep{chen2026dflash}: block diffusion speculative decoding with a single sequence draft.
  \item \textbf{DDTree-oracle} \citep{ringel2026accelerating}: DDTree with the best budget per (dataset, model, temperature),
        selected by grid search over $\{16, 32, 64, 128, 256, 512, 1024\}$.
\end{itemize}

\paragraph{Hardware and hyperparameters.}
All experiments run on 8 A800 GPUs with block size $L = 16$, \texttt{bfloat16} precision, maximum 2048 new tokens, and temperatures $\{0.0, 1.0\}$.
For \caDDTree{}, cost parameters are estimated once per (model, hardware) pair via the profiling procedure in Section~\ref{sec:method-cost-estimation}.
For DDTree and \caDDTree{}, the target model uses standard PyTorch scaled dot product attention, as the standard FlashAttention-2 \citep{dao2024flashattention} kernel does not support the custom tree attention mask; the DFlash drafter uses FlashAttention-2.
For the AR and DFlash baselines, we report the faster of the two attention implementations.

\subsection{Main Results}
\label{sec:exp-main}

Table~\ref{tab:main-results} reports per-token generation time (ms/tok; lower corresponds to higher throughput) and mean acceptance length $\tau$ at temperature~0.0.
Temperature~1.0 results are in Appendix~\ref{sec:appendix-results}.

\caDDTree{} matches or surpasses DDTree-oracle on nearly all tasks.
For Qwen3-8B, the grid oracle is constrained to $B{=}128$; \caDDTree{} selects budgets between grid values and outperforms the oracle on most tasks.
For Qwen3-4B, the oracle uses $B{=}256$; \caDDTree{} closely matches the oracle across tasks without any grid search.
The pattern is consistent at both temperatures (temperature-1.0 results in Appendix~\ref{sec:appendix-results}).

\subsection{Cost Model Validation}
\label{sec:exp-cost-model}

Figure~\ref{fig:cost-profile} shows measured verification latency against node count for nine context lengths (0--8k tokens).
The measured latency is consistent with being monotone and convex in node count across all contexts; the convex regression fits closely track the measurements ($R^2 \geq 0.979$ across all nine context lengths, RMSE\,$\leq$\,4.9\,ms), supporting Assumption~\ref{ass:convex}.
Moreover, both the base verification cost $C_v(0;\,\ell)$ and the marginal cost per node increase with context length; by design, this causes \caDDTree{} to select smaller budgets at longer contexts without any manual adjustment.
Table~\ref{tab:cost-model-ablation} compares two cost model variants.

\begin{table}[tbp]
\centering
\small
\setlength{\tabcolsep}{5pt}
\caption{Cost model ablation (Qwen3-4B, temp.~0.0). Per-token latency in ms/tok; lower is better. ``Constant'' sets $C_v(n;\,\ell){=}C_v(0;\,\ell)$ for all~$n$, causing the algorithm to always expand to the maximum budget; latency increases about $2{\times}$ across tasks.}
\label{tab:cost-model-ablation}
\begin{tabular}{lccc}
\toprule
$C_v$ model & MATH-500 & AIME 2025 & Alpaca \\
\midrule
\caDDTree{}  & \textbf{4.53} & \textbf{4.96} & \textbf{10.66} \\
Constant     & 9.14          & 10.25         & 19.37 \\
\bottomrule
\end{tabular}
\end{table}

The constant model sets $C_v(n;\,\ell) = C_v(0;\,\ell)$ for all $n$, so the surrogate throughput is strictly increasing and the algorithm always grows the tree to the maximum budget (capped at 1024 nodes), yielding a clear per-token latency increase and confirming that per-node verification overhead is non-negligible.

\begin{figure*}[tp]
\centering
\includegraphics[width=\textwidth]{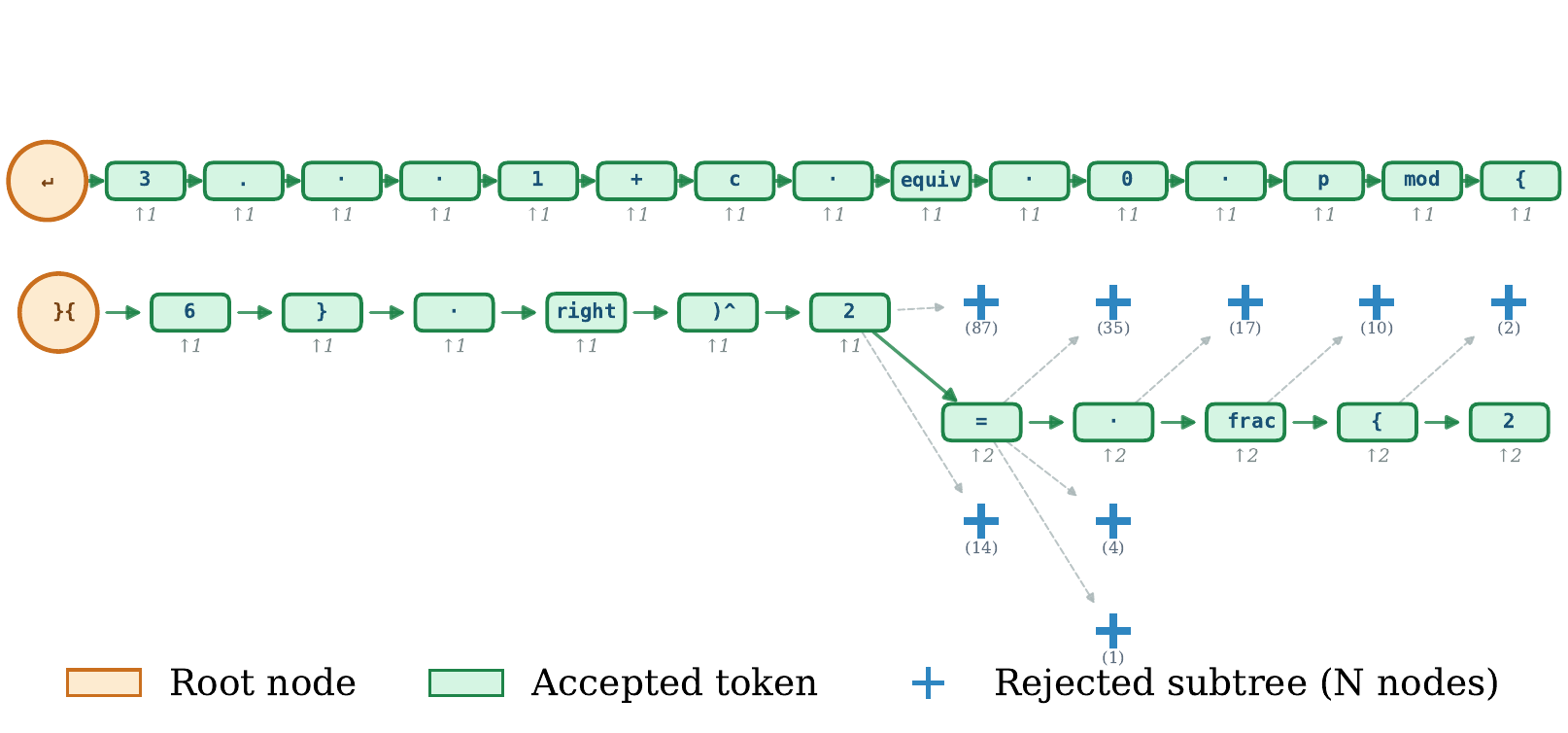}
\caption{Throughput-optimal draft trees constructed by \caDDTree{} for two decoding rounds (MATH-500, Qwen3-8B, temp.\,0.0), showing the rank of each accepted token among its sibling candidates.
The orange circle is the bonus token (tree root); green boxes are accepted tokens labeled with their rank ($\uparrow\!k$: the accepted token was the $k$-th most probable child); a \textbf{+} marker with count $N$ below it indicates a rejected subtree of $N$ nodes.
\emph{Top ($n^*\!=\!15$, 15 tokens accepted):} all accepted tokens are rank-1 with no competing siblings; the drafter is near-certain and a narrow chain suffices, saving 87\% of the oracle budget.
\emph{Bottom ($n^*\!=\!181$, 13 tokens accepted):} accepted tokens span multiple ranks and large rejected subtrees appear at every step; wide exploration is necessary to find the accepted path.
The two regimes coexist within the same dataset; no fixed budget serves both, and \caDDTree{} selects $n^*$ automatically each round.}
\label{fig:tree-example}
\end{figure*}

\subsection{Adaptive Budget Analysis}
\label{sec:exp-adaptive}

The throughput-optimal $n^*$ varies substantially across rounds, even within the same dataset.
Figure~\ref{fig:adaptive-budget} makes this concrete: on MATH-500, budgets spread widely, with a substantial fraction of rounds selecting fewer nodes than the grid oracle and a larger fraction selecting more.
No single fixed budget serves both regimes well.
On Alpaca, the distribution is tightly concentrated and the mean falls between grid points, at a value no fixed grid budget matches exactly.
\caDDTree{} finds the per-round optimum automatically in both cases.

Figure~\ref{fig:tree-example} reveals \emph{why} the optimal budget differs: the drafter's per-token uncertainty varies across steps.
When the drafter is confident, every accepted token sits at rank~1 with no competing siblings; a narrow chain suffices, saving most of the oracle budget at no loss in accepted tokens.
When the drafter is uncertain, accepted tokens span multiple ranks and large rejected subtrees appear at every step; the tree must fan out widely to find the accepted path.
Together, the two panels confirm that per-round budget adaptation captures both regimes, and that no single fixed budget can eliminate this variation.

\subsection{Ablation: Sensitivity to Cost Parameters}
\label{sec:exp-sensitivity}

\begin{table}[tbp]
\centering
\small
\setlength{\tabcolsep}{5pt}
\caption{Sensitivity of \caDDTree{} per-token latency to cost parameter perturbations (Qwen3-4B, MATH-500, temp.~0.0). $\Delta$ is the change relative to the nominal (unperturbed) run.}
\label{tab:sensitivity}
\begin{tabular*}{\columnwidth}{@{\extracolsep{\fill}}lcccc}
\toprule
 & $C_d{\times}0.5$ & $C_d{\times}2.0$ & $C_v{\times}0.5$ & $C_v{\times}2.0$ \\
\midrule
$\Delta$ (ms/tok) & $-0.07$ & $+0.05$ & $-0.04$ & $-0.01$ \\
\bottomrule
\end{tabular*}
\end{table}

Table~\ref{tab:sensitivity} shows that all four perturbations produce $|\Delta| \leq 0.07$\,ms/tok, suggesting that approximate cost estimates suffice and highly precise offline calibration is not strictly necessary.

\section{Conclusion}
\label{sec:conclusion}

We introduced \caDDTree{}, a cost-aware diffusion draft tree method that optimizes per-token latency directly rather than acceptance length under a fixed budget.
By explicitly modeling both drafting and verification latencies, we showed that the throughput objective decomposes into a one-dimensional budget search that is efficiently solvable via a greedy stopping rule with provable unimodality guarantees under convex verification cost.
\caDDTree{} eliminates offline budget selection and adapts the tree size per decoding round, consistently matching or exceeding DDTree with oracle budget selection across diverse benchmarks and target models.

\clearpage
\bibliography{custom}

\clearpage
\appendix
\section*{Appendix}

\section{Proofs}
\label{sec:proofs}

\subsection{Derivation of Per-Round Objective}
\label{sec:proof-throughput}

Over $R$ rounds, the response throughput is $\Theta_R = \sum_{t=1}^R (1+\alpha_t) / \sum_{t=1}^R C(\mathcal{T}_t;\,\ell_t)$.
At each round, $\mathcal{T}_t$ is a deterministic function of the drafter's marginals $\{q_i^{(t)}\}$, while acceptance $\alpha_t$ is random under the target $p$.
By the law of large numbers, as $R \to \infty$:
\begin{align*}
  \Theta_R \to \Theta
  = \frac{\mathbb{E}_t[\mathbb{E}_p[1+\alpha_t \mid \mathcal{T}_t]]}
         {\mathbb{E}_t[C(\mathcal{T}_t;\,\ell_t)]}
  \approx \frac{\mathbb{E}_t[1 + \Phi(\mathcal{T}_t)]}
               {\mathbb{E}_t[C(\mathcal{T}_t;\,\ell_t)]},
\end{align*}
where the inner expectation over target acceptance is replaced by the surrogate $\Phi$.
We therefore maximize $\hat{\theta}(\mathcal{T}_t;\,\ell_t)$ at each round as a tractable per-round surrogate for the long-run ratio $\Theta$.
\hfill$\square$

\subsection{Proof of Proposition~\ref{prop:decomposition}}
\label{sec:proof-decomp}

\textbf{Fixed-budget optimality.}
For any node $\mathbf{s} = (s_1, \ldots, s_d)$, its parent satisfies
$\pi_{\mathbf{s}_{-}} = \pi_{\mathbf{s}} / q_d(s_d) \ge \pi_{\mathbf{s}}$,
so every ancestor has weakly higher prefix probability.
Hence if $\mathbf{s}$ is among the top-$n$ nodes by $\pi$, every ancestor is too:
$\mathcal{T}^*_n$ is prefix-closed and forms a valid tree.
Since $\Phi(\mathcal{T}) = \sum_{\mathbf{s}\in\mathcal{T}}\pi_{\mathbf{s}}$ and
$\mathcal{T}^*_n$ selects the $n$ nodes with the largest $\pi_{\mathbf{s}}$,
no valid $n$-node tree achieves a higher $\Phi$.

\textbf{Budget selection.}
For fixed $n$, the denominator $C_d + C_v(n;\,\ell)$ is constant, so maximizing $\hat{\theta}(\mathcal{T};\,\ell)$
over valid $n$-node trees reduces to maximizing $\Phi(\mathcal{T})$, achieved by $\mathcal{T}^*_n$.
Taking the maximum over $n \ge 0$ yields~\eqref{eq:decomposition}.
\hfill$\square$

\subsection{Proof of Lemma~\ref{lem:weighted-avg}}
\label{sec:proof-lemma}

We verify the weighted average representation by direct expansion.
Let $c_{n+1} = C_v(n+1;\,\ell) - C_v(n;\,\ell) > 0$ and $w_n = (C_d + C_v(n;\,\ell)) / (C_d + C_v(n+1;\,\ell))$,
so $1 - w_n = c_{n+1} / (C_d + C_v(n+1;\,\ell))$.
Then:
\begin{align}
  &w_n \hat{\theta}^*(n) + (1-w_n)\,\rho_{n+1} \nonumber\\
  &= \frac{1 + \Phi^*(n)}{C_d + C_v(n+1;\,\ell)}
   + \frac{c_{n+1}}{C_d + C_v(n+1;\,\ell)}
     \cdot \frac{\delta_{n+1}}{c_{n+1}} \nonumber\\
  &= \frac{1 + \Phi^*(n) + \delta_{n+1}}{C_d + C_v(n+1;\,\ell)}
   = \hat{\theta}^*(n+1). \nonumber
\end{align}

The equivalence $\hat{\theta}^*(n+1) > \hat{\theta}^*(n)$ iff $\rho_{n+1} > \hat{\theta}^*(n)$
follows immediately: $\hat{\theta}^*(n+1)$ is a strict convex combination of $\hat{\theta}^*(n)$
and $\rho_{n+1}$, so it lies between them, and exceeds $\hat{\theta}^*(n)$ iff
$\rho_{n+1} > \hat{\theta}^*(n)$.
\hfill$\square$

\subsection{Proof of Theorem~\ref{thm:unimodal} (Unimodality)}
\label{sec:proof-thm}

By Assumption~\ref{ass:convex}, $\{\rho_n\}$ is non-increasing.
If $\hat{\theta}^*$ is non-decreasing for all $n$, the conclusion holds trivially.
Otherwise, let $n_0 \ge 0$ be the smallest index with $\hat{\theta}^*(n_0+1) \le \hat{\theta}^*(n_0)$,
equivalently $\rho_{n_0+1} \le \hat{\theta}^*(n_0)$.
From the weighted average representation:
\begin{align}
  \hat{\theta}^*(n_0+1)
  &= w_{n_0}\hat{\theta}^*(n_0) + (1-w_{n_0})\rho_{n_0+1} \nonumber\\
  &\ge \rho_{n_0+1}.
  \label{eq:lb-app}
\end{align}

\textbf{Inductive step.}
Assume $\rho_{n+1} \le \hat{\theta}^*(n)$ for some $n \ge n_0$,
so $\hat{\theta}^*(n+1) \le \hat{\theta}^*(n)$ and the bound above gives
$\hat{\theta}^*(n+1) \ge \rho_{n+1}$.
Since $\{\rho_n\}$ is non-increasing:
\[
  \rho_{n+2} \le \rho_{n+1} \le \hat{\theta}^*(n+1),
\]
so $\hat{\theta}^*(n+2) \le \hat{\theta}^*(n+1)$ by Eq.~\eqref{eq:stopping}.
The induction is complete: $\hat{\theta}^*$ is non-increasing for all $n \ge n_0$.
\hfill$\square$

\subsection{Proof of Corollary~\ref{cor:greedy} (Greedy Optimality)}
\label{sec:proof-cor}

Algorithm~\ref{alg:caddtree} terminates at the first $n$ for which
$\hat{\theta}^*(n+1) \le \hat{\theta}^*(n)$.
By Theorem~\ref{thm:unimodal}, this is the global maximizer $n^*$.
At termination, the heap has popped exactly the top-$n^*$ prefixes by probability
(by Proposition~\ref{prop:decomposition}),
and Proposition~\ref{prop:decomposition} guarantees that $\mathcal{T}^*_{n^*}$ is the
throughput-optimal tree.
\hfill$\square$

\subsection{Proof of Corollary~\ref{cor:convex} (Convex Cost)}
\label{sec:proof-convex}

$\{\delta_n\}$ is non-increasing by construction; $\{c_n\}$ non-decreasing by the convexity
hypothesis; hence $\rho_n = \delta_n / c_n$ is non-increasing.
\hfill$\square$

\section{Benchmark Details}
\label{sec:appendix-benchmark}

All runs use block size 16, \texttt{bfloat16} precision, and a maximum of 2048 new tokens.
Each dataset uses 128 evaluated examples, except AIME 2025 (30), HumanEval (164), and MT-Bench (160).
The evaluation set is sharded across 8 A800 GPUs.
Before timing, one warmup prompt is run through each method (autoregressive, DFlash, DDTree,
\caDDTree{}) to exclude one-time setup costs.

\section{Temperature 1.0 Results}
\label{sec:appendix-results}

\begin{table*}[h]
\centering
\small
\setlength{\tabcolsep}{4pt}
\caption{Per-token generation time (ms/tok) and mean acceptance length $\tau$ at temperature~1.0. $n^*$ is the oracle grid budget (same as T=0 grid search); $\bar{n}^*(\sigma)$ is \caDDTree{}'s per-round mean (std). \caDDTree{} achieves the lowest per-token latency without offline budget search.}
\label{tab:main-results-temp1}
\begin{tabular*}{\linewidth}{@{\extracolsep{\fill}}l c cc ccc ccc}
\toprule
 & \textbf{AR}
 & \multicolumn{2}{c}{\textbf{DFlash}}
 & \multicolumn{3}{c}{\textbf{DDTree-oracle}}
 & \multicolumn{3}{c}{\textbf{\caDDTree{} (ours)}} \\
\cmidrule(lr){3-4}\cmidrule(lr){5-7}\cmidrule(lr){8-10}
\textbf{Dataset}
  & ms/tok
  & ms/tok & $\tau$
  & ms/tok & $\tau$ & $n^*$
  & ms/tok & $\tau$ & $\bar{n}^*(\sigma)$ \\
\midrule
\multicolumn{10}{c}{\emph{Qwen3-4B}} \\
MATH-500       & 30.76 & 6.95 & 6.68 & \textbf{5.17} & 9.04  & 256 & 5.20          & 9.26 & $224\,(51)$ \\
GSM8K          & 31.05 & 7.64 & 6.00 & \textbf{5.50} & 8.65  & 256 & 5.56          & 8.67 & $250\,(59)$ \\
AIME 2025      & 30.60 & 9.17 & 5.04 & \textbf{6.34} & 7.27  & 128 & 6.58          & 7.39 & $220\,(31)$ \\
HumanEval      & 31.01 & 7.61 & 5.94 & \textbf{5.44} & 8.73  & 256 & 5.49          & 8.64 & $227\,(55)$ \\
MBPP           & 30.79 & 8.08 & 5.66 & \textbf{5.64} & 8.43  & 256 & 5.80          & 8.32 & $260\,(57)$ \\
LiveCodeBench  & 30.70 & 7.06 & 6.61 & 5.36          & 8.75  & 128 & \textbf{5.35} & 9.05 & $215\,(38)$ \\
MT-Bench       & 31.09 & 13.53 & 4.01 & 9.02         & 5.84  & 256 & \textbf{8.94} & 6.05 & $235\,(36)$ \\
Alpaca         & 30.66 & 16.71 & 2.98 & \textbf{10.96} & 4.63 & 256 & 11.20        & 4.63 & $261\,(40)$ \\
\midrule
\multicolumn{10}{c}{\emph{Qwen3-8B}} \\
MATH-500       & 31.10 & 7.31 & 6.46 & 5.48          & 8.73  & 128 & \textbf{5.40} & 9.02 & $168\,(36)$ \\
GSM8K          & 30.84 & 7.69 & 5.92 & 5.67          & 8.27  & 128 & \textbf{5.62} & 8.44 & $184\,(23)$ \\
AIME 2025      & 31.05 & 9.65 & 4.94 & 6.98          & 6.94  & 128 & \textbf{6.80} & 7.11 & $158\,(33)$ \\
HumanEval      & 31.73 & 8.33 & 5.54 & \textbf{5.93} & 7.99  & 128 & 5.96          & 8.11 & $174\,(34)$ \\
MBPP           & 31.60 & 8.95 & 5.21 & 6.27          & 7.62  & 128 & \textbf{6.18} & 7.89 & $187\,(18)$ \\
LiveCodeBench  & 30.83 & 6.96 & 6.84 & 5.30          & 9.02  & 128 & \textbf{5.29} & 9.24 & $162\,(33)$ \\
MT-Bench       & 30.92 & 14.46 & 3.78 & 9.68         & 5.54  & 128 & \textbf{9.47} & 5.77 & $169\,(28)$ \\
Alpaca         & 31.67 & 17.47 & 2.94 & 11.64        & 4.44  & 128 & \textbf{11.55} & 4.54 & $186\,(11)$ \\
\bottomrule
\end{tabular*}
\end{table*}

\end{document}